\documentclass[10pt, a4paper]{article}
\usepackage{lrec}
\usepackage{multibib}
\newcites{languageresource}{Language Resources}
\usepackage{graphicx}
\usepackage{tabularx}
\usepackage{soul}
\usepackage{bbm}
\usepackage{multirow}

\usepackage{epstopdf}
\usepackage{amsmath}
\usepackage{booktabs}
\usepackage{arydshln}
\usepackage[latin1]{inputenc}
\usepackage{url}
\usepackage{hyperref}
\usepackage{xstring}
\usepackage{diagbox}
\usepackage{subcaption}
\usepackage{booktabs}
\title{Evaluating Text Coherence at Sentence and Paragraph Levels}

\name{Sennan Liu$^{13}$, Shuang Zeng$^{23}$, Sujian Li$^{3}$}
\address{$^{1}$Department of English and International Study, China Foreign Affairs University, Beijing, China\\
$^{2}$School of Software and Microelectronics, Peking University, Beijing, China\\
$^{3}$MOE Key Lab of Computational Linguistics, Peking University, Beijing, China\\
         \url{stan.x.liu.18@gmail.com}, \url{{zengs, lisujian}@pku.edu.cn} }

\abstract{
In this paper, to evaluate text coherence, we propose the paragraph ordering task as well as conducting sentence ordering. 
We collected four distinct corpora from different domains on which we investigate the adaptation of existing sentence ordering methods to a  paragraph ordering task. 
We also compare the learnability and robustness of existing models by artificially creating mini datasets and noisy datasets respectively and verifying the efficiency of established models under these circumstances.
Furthermore, we carry out human evaluation on the rearranged passages from two competitive models and confirm that WLCS-l is a better metric performing significantly higher correlations with human rating than $\tau$, the most prevalent metric used before. 
Results from these evaluations show that except for certain extreme conditions, the recurrent graph neural network-based model is an optimal choice for coherence modeling. \\ \newline  }

\begin{document}
\maketitleabstract

\section{Introduction}
\label{sec:introduction}

Coherence modeling has been a topic for discourse analysis for a long time~\cite{lapata-2003-probabilistic}. 
Text ordering is the standard task used to test a coherence model in NLP.
% , and is essential for many downstream NLP tasks~\cite{barzilay-mcKeown-2005-summerization,hovy-1988-planning,verberne-2007-q&a}.
While earlier work aims at distinguishing between a coherently ordered list of sentences and a random permutation thereof, recent studies attempt to generate a consecutive text from a set of given sentences. 
% Two reasons probably contributes to the shift: discriminative models are prone to overfitting on the particular dataset and domain, and the combinatorial explosion brought by $\n!$ nature of random permutation.

Various frameworks exist, focusing on linguistic features via statistical models ~\cite{lapata-2003-probabilistic,barzilay-lee-2004-catching,barzilay-lapata-2005-entity,elsner-charniak-2011-extending}. Especially, entity based models ~\cite{barzilay-lapata-2008-modeling,guinaudeau-2013-graphbased} have shown the effectiveness of exploiting entities for this task. 
% Sequence ordering is one way to formulate coherence modeling problems, in which a reasonable order is supposed to be generated from a set of input sequences. Intuitively, the basic idea is to use an higher-level encoder or discriminator to figure out the relationship between each of the bottom level encoded sequences.
% Measures to do sequence encoding evolves from human created features to sequential neural network models.
% Intuitively, the basic idea is to use an higher-level encoder or discriminator to figure out the relationship between each of the bottom level encoded sequences.
% Measures to do sequence encoding evolves from human created features to sequential neural network models.
% Earlier works on sequence ordering mainly focused on introducing of linguistic features via statistical models ~\cite{lapata-2003-probabilistic,barzilay-lee-2004-catching,barzilay-lapata-2005-entity,elsner-charniak-2011-extending}. 
% Especially, entity based models ~\cite{barzilay-lapata-2008-modeling,guinaudeau-2013-graphbased} have shown the effectiveness of exploiting entities for this task. 
% Later, studies have evolved neural network into models,
With the popularization of neural networks, studies
formulate the problem as neural pair-wise discrimination~\cite{li-hovy-2014-model,chen-2016-neuralSO}, with a purpose on training a discriminator to separate coherent clique or pairs from non-coherent ones. 
% Later, sentence ordering tesk~\cite{li-jurafsky-2017-neural}
Recently, thanks to the success of set-to-sequence framework ~\cite{vinyals-2015-ptrnet,vinyals-2016-ordernet}, the focus shifts to generative ordering problem  ~\cite{gong-2016-endtoendNS,logeswaran-2017-sentenceOU,cui-etal-2018-deep,yin-2019-grn}, aiming at finding a framework to learn to generate an optimal order from a bunch of input sequences. 
However, most of these studies only focus on the sentence level task and experiment using the abstracts from academic papers, which is unrealistic in the real world where various lengths and text domains may be used. 
Moreover, none of them analyze the adaptability of models with respect to different qualities of language resource, which prevents further study to improve the models in a fine-grained aspect.

In this paper, we firstly propose  to study text ordering at the paragraph level as a supplement of sentence ordering, with a purpose of comprehensively researching text coherence.
We then compare the performance of the existing models on 4 prevalent sentence-level datasets and 4 paragraph-level datasets from diverse domains we collected.
Then we design artificial datasets to evaluate their learnability and robustness by shrinking training size and adding artificial noise to the language source respectively, which leads to several conclusions that could contribute to further studies. 
Finally, we carry out human evaluation to assess coherence of machine-arranged text and find WLCS-l a much more reliable metric on sentence ordering task than the most prevailing metric before. 
We believe our discoveries will guide further research and application.
\section{Datasets}
\label{sec:data}
In this section, we will briefly go through the prevalent datasets and introduce the corpus we compiled.
%, including the mini datasets and noisy datasets we created for learnability and robustness evaluation. 

In order to fairly verify performance of the major approaches at both the sentence and the paragraph level, 8 datasets are used in total, among which 4 are at the sentence level and the rest are at the paragraph level.
% We also pick and process one datasets from each level for the purpose of testing the learnability and robustness.

\begin{table*}[ht]
\begin{center}
\setlength{\tabcolsep}{1.6mm}{
\begin{tabular}{ccccccccccc}
\toprule
\multirow{2}{*}{Dataset} & \multicolumn{3}{c}{Average sequence number}                                    & \multicolumn{3}{c}{Average sequence length}                                    & \multicolumn{3}{c}{Data split}                                                 & \multirow{2}{*}{Vocabulary} \\ \cline{2-10}
                         & \multicolumn{1}{c}{Train} & \multicolumn{1}{c}{Dev} & \multicolumn{1}{c}{Test} & \multicolumn{1}{c}{Train} & \multicolumn{1}{c}{Dev} & \multicolumn{1}{c}{Test} & \multicolumn{1}{c}{Train} & \multicolumn{1}{c}{Dev} & \multicolumn{1}{c}{Test} &                             \\ \hline
 NIPS & 6.40 &  6.66  &  6.48  & 25.92 & 26.93  &  26.40  &  3519  &  196  &  195 &  19957 \\  
  AAN & 5.08 &  5.31  & 5.51 & 24.56 & 24.51  &  24.47  & 15961 & 892 & 893 & 55588 \\
  arXiv & 6.69 & 6.98 & 7.23 & 25.19 & 25.69 & 25.17  & 36652 & 2045  &  2044 & 97305 \\
  SIND & 5.00 &  5.00  & 5.00 & 11.49 & 11.62  & 11.42  & 40155 & 4990 &  5055  & 30952 \\ \hline
  News & 7.97(3.16) &  7.96(3.16)  & 8.07(3.15) & 77.45 & 77.46  & 75.78  & 12634 & 720 & 707 & 87067 \\
  Statements & 5.58(3.23) & 5.47(3.54)  & 4.85(3.65) & 66.30 & 86.99 & 85.76  & 12530 & 712 & 715 & 53708 \\
  Economist & 9.73(4.31) &  10.04(4.69)  & 9.94(4.78) & 94.81 & 97.53  & 97.36  &  70728 & 3929 &  3929  & 535345 \\
  Lyrics & 29.51(2.21) & 29.42(2.23)  & 29.51(2.25) & 7.28 & 7.27  &  7.27  &  43858  &  2436 & 2436 & 77021 \\
\bottomrule
\end{tabular}}
\caption{Statistics of eight datasets used in our experiments. For datasets at the paragraph level, we count the average number of sentences in each paragraph and show them in the parenthesis. %Note that we didn't count paragraph with only one sentence. This could be a reflection of the average informativeness of a paragraph in comparison with sentence ordering.
}
\label{tab:ds-stats}
\end{center}
\end{table*}

\subsection{Datasets for sentence ordering}
Four existing datasets are adopted to evaluate model performance at the sentence level. 
\begin{enumerate}
    \item \textbf{NIPS Abstract}. This dataset contains roughly 3K abstracts from NIPS papers from 2005 to 2015.\footnote{\url{https://www.kaggle.com/benhamner/nips-papers}}.
    \item \textbf{ANN Abstract}. It includes about 13K abstracts extracted from the papers in the ACL Anthology Network (AAN) corpus\cite{radev-2016-AAN}. 
    \item \textbf{arXiv Abstract}. We further consider another source of abstracts collected from arXiv. It consists of around 40k instances\footnote{\url{https://www.kaggle.com/neelshah18/arxivdataset}}.
    \item \textbf{SIND}. It has 50K stories for the visual storytelling task\cite{huang-etal-2016-visual}, which are from a different domain. 
    %Here we use each story as a paragraph.
\end{enumerate}

Statistics of the datasets are listed in Table~\ref{tab:ds-stats}.

\subsection{Datasets for document ordering}

To assess existing approaches at the paragraph level and in broader domains, we collect four datasets for paragraph ordering. 
\begin{enumerate}
    \item \textbf{News}. This dataset consists of news releases of U.S. Department of Justice(DOJ), from January 2009 to August 2019. 
    We create the corpus by first crawling data\footnote{Scraper repo: \url{https://github.com/jbencina/dojreleases}.
    We removed the speeches when scraping.} from the DOJ website and then parse and filter the disqualified articles. 
    For the bullet points or numbering lists, we link them to the tail of the previous paragraph successively.
    %In this way the units in one paragraph is much more integrated and the total length of one certain passage is compressed without its coherence being hurt. 
    \item \textbf{Statements}. This dataset contains American presidential written statements. %from the birth of United States. 
    The data source is from the website of UCSB presidency project\footnote{ \url{https://www.presidency.ucsb.edu/documents/app-categories/presidential/written-statements}}. 
    We integrate all the bullet points and numbering lists with their starting paragraph.
    \item \textbf{Economist}. This dataset contains the articles from the famous journal Economist from 1990 to 2018. Ads are removed with only texts left.\footnote{ \url{https://tea.share2china.com/}.
    %Note: The source is not publicly available. We brought the copy right from its owner for only the research purpose.
    }
    \item \textbf{Lyrics}. This dataset is made of the lyrics of pop songs\footnote{\url{https://www.kaggle.com/mousehead/songlyrics}}. 
    Each line of lyrics is perceived as a sequence in our experiments.
    % ################################
    Although they are not strictly paragraphs, the weak connect between each sequence makes it hard to find a logical order as sentence ordering.
    % ################################
\end{enumerate}

Our datasets differ from the previous ones in two ways. 
First, each sequence is less connected semantically and there are not as much conjunctive tokens as hint as in abstracts.
Second, they are from diverse domains, resulting in a good chance to examine the adaptability of models assessed in different themes.
Each dataset we collected is split in the same ratios - 90\% for training and  each 5\% for development and testing. 
Except for \textbf{Lyrics},  we split all other corpus in a chronological order, which means the models are trained and tuned on the text released in an earlier time and tested on the ones released afterward. 
The statistics of the four are displayed in Table~\ref{tab:ds-stats}.

\section{Method}
\label{sec:method}
In this work, to evaluate the performance of text ordering,
we choose four state-of-the-art generative models which outperform earlier discriminative models.
%, we  introduce the methods  
%from the perspective of models. 
%we choose generative models in our evaluation including the state-of-the-art approach for their proved excellent performance compared with earlier discriminative models. 
%In total, four models are involved in our evaluations, 
These four models are:  (1)\textbf{LSTM+Ptr-Net}~\cite{gong-2016-endtoendNS}, (2)\textbf{Variant-LSTM+Ptr-Net}~\cite{logeswaran-2017-sentenceOU}, (3)\textbf{ATTOrderNet}~\cite{cui-etal-2018-deep}, (4) \textbf{SE-Graph}~\cite{yin-2019-grn}. 
Their decoders are all based on the Ptr-Net framework~\cite{vinyals-2015-ptrnet}.% as the decoder.  
Through the use of attention~\cite{bahdanau-2014-neuralMT,luong-2015-effectiveAT}, Ptr-Net works as a type of new decoder for sequence generation. 

While a normal attention network picks the weighted average of all vectors (e.g. word embeddings of a sentence) as the input for the current time step, Ptr-Net simply uses the attention value as the pointer and picks out one tensor from all input tensors. 
In this way, attention is used to retrieve the most likely vector as input for RNN and as a probability index when generating the output pointer.
Formally,  Ptr-Net can be described as: 
\begin{align}
    h_{dec}^{t}, c_{dec}^{t} &= LSTM\big(h_{dec}^{t-1}, c_{dec}^{t-1}, x^{t-1}\big) \\
    e_{dec}^{t,i} &= f\big(s_{i}, h_{dec}^{t}\big); i \in \{1,..., n\} \\
    a_{dec}^{t} &= Softmax\big(e_{dec}^{t}\big)
\end{align}
% The LSTM takes the embedding of the previous sentence as input instead of the attention readout. 
% At training time the correct order of sentences $(s_{o1} , s_{o2} , ..., s_{on} ) = (x_{1}, x_{2}, ..., x_{n})$ is known ($o$ represents the correct order) and $x^{t-1}$ is used as the input.
% At test time the predicted assignment $x^{t-1}$ is used instead. The attention
% computation is identical to that of the encoder, but now $a_{dec}^{t,i}$ is interpreted as the probability for $s_{i}$ being the correct sentence choice at position t, conditioned on the previous sentence assignments $p(S_{t} = s_{i}|S_{1}, ..., S_{t-1})$. 
% The initial state of the decoder LSTM is initialized with the final hidden state of the encoder. 
% $x_{0}$ is a vector of zeros.

All the models based on a hierarchical architecture~\cite{yang-etal-2016-hierarchical}, in which  LSTM~\cite{hochreiter-1997-lstm} works as the bottom level sequence encoder, encoding the sentence into a compressed representation in a way it can fit on the upper level structure for reasoning. 

The major differences of these models lie in their encoders. {\it{\textbf{LSTM+PtrNet}}} uses a conventional LSTM to read the paragraphs or sentences representation and learn the representations as a whole.
{\it{\textbf{Variant-LSTM+PtrNet}}} is based on the set-to-sequence framework~\cite{vinyals-2016-ordernet}, which reads the discourse by a series of RNN units with weighted average of sequences vectors as input for each time step and passes the last hidden layer to decoder as the final high level abstraction. 
{\it{\textbf{ATTOrderNet}}} adopts self-attention architecture~\cite{vaswani-2017-attention},where the input is the concatenated sequence vectors for a self-attention encoder without the position encoding information. 
{\it{\textbf{SE-Graph}}} utilizes the Graph Recurrent Network~\cite{zhang-2018-sentencestateLF}, that parallelly and iteratively updates its node states with a message passing framework~\cite{gilmer-2017-neuralMP}. 
For every message passing step $t$, the state of each node update involves a massage calculating from its directly connected neighbors and applying the gated operations with the newly calculated message.

\section{Experiments and Results}
\label{sec:experiments}
In this section, we will elaborate our experiment settings and discuss our results. We first use our newly assembled sources to evaluate the models across different sequence levels and genres. Then we make a further assessment on robustness under diverse data scale and noisy extent. Finally, human evaluation is carried out to compare the correlation between human rating and several existing metrics.

\subsection{Experiment Setting}
 For fair comparison, all the batch sizes share the size of 8 and beam width of 32. All of of the models use LSTM as sequence encoder. We use early stopping when there is no performance improvement in 5 consecutive epochs. The nltk tokenizer is used for word tokenization\footnote{NLTK implementation: \url{http://www.nltk.org/}}. All the out-of-vocabulary words are replaced with $<$UNK$>$, whose embeddings are adapted during the training process.
For \textbf{LSTM+PtrNet} and \textbf{SE-Graph}, the 100 dimension GloVe 6B~\cite{pennington-etal-2014-glove} pretrained word embedding vectors are used to initialize word embedding layer, while for \textbf{Variant-PtrNet} and \textbf{ATTOrderNet}, we use GloVe 300d instead.
Due to the space limitation, we only emphasize some of the crucial settings. For each model, the specific configurations used are listed as follows: 

\begin{itemize}

    \item For \textbf{LSTM+PtrNet}, we follow the authentic configuration and a scoring function for pointer generation is set to be an generalized version of global attention~\cite{bahdanau-2014-neuralMT}:
    \begin{equation}
        f(s,h) = W'(W[s;h] + b) + b'\\
    \end{equation}
    In order to explore the potential of this framework, we employ 4 parallel heads throughout all pointer attention layers\footnote{As we fail to aquire the source code from the author, we replicate the model by ourselves. In practice we found the performance of our version approximates to the authentic release.} and use L2 weight decay on the trainable parameters with regularization parameter $\lambda = 10^{-5}$.
    
    \item For \textbf{Variant-LSTM+PtrNet} we employ 8 parallel heads, like in \textbf{LSTM+PtrNet}, throughout all pointer attention layers. All other hypterparameters follow the setting of previous work. \footnote{As we tried our best but still cannot get connection to the author, we replicate the model by ourselves. Some performance mismatch are observed, but the framework completely follows the original paper.}.
    % , the hidden size of all LSTM architecture is set to 1000, and the scoring function for pointer generation is set to be an adapted version of local attention~\cite{luong-2015-effectiveAT}:
    % \begin{equation}
    %     f(s,h) = s^{T}(W h + b)\\
    % \end{equation}
    % Adam~\cite{kingma-2014-adam} is used as the optimizer. The learning rate is initialized to $10^{-5}$,
    
    \item For \textbf{ATTOrderNet}, the number of attention hidden layers is {6, 4, 4, 2} for AAN abstract, arxiv abstract, Economist and the rest, respectively. We employ 8 parallel heads throughout all self-attention layers. Any other settings keep the same as the original paper.
    % Adadelta~\cite{zeiler-2012-adadelta} 
    % Settings related to weight decay and regularization keeps the same as those of \textbf{LSTM+PtrNet}
    % was adopted as the optimizer with $\epsilon = 10^{-6}$ and $\rho = 0.95$. The learning rate is initialized to 1.0. The hidden layer size of LSTMs in sentence encoder is 256, and is 512 in the decoder. 
    
    \item For \textbf{SE-Graph}, the setting of the optimizer is the same as the setting of \textbf{ATTOrderNet}. We use the hidden size of 300 for both sequence encoder and GRN encoders for all sentence-level datasets except SIND. Besides these configurations, we keep all the rest hyperparameters as what is in the original release.
\end{itemize}
Except for SE-Graph, which is implemented by Pytorch, all of the models are implemented with TensorFlow.
\begin{table*}[ht]
\begin{center}
\setlength{\tabcolsep}{1.39mm}{
\begin{tabular}{ccccccccccccc}
\toprule
\multirow{2}{1.5cm}{\centering Sentence Level}  & \multicolumn{3}{c}{arXiv}                                     & \multicolumn{3}{c}{AAN}                                  & \multicolumn{3}{c}{NIPS} 
                            & \multicolumn{3}{c}{SIND} \\ \cline{2-13}
                         & pmr & wlcs-l & \#pm & pmr & wlcs-l & \#pm & pmr & wlcs-l & \#pm & pmr & wlcs-l & \#pm \\ \midrule
LSTM+PtrNet & 22.17 & 0.6002 & 11.0M & 35.65 & 0.6575 & 6.8M & 19.79 & 0.5839 & 3.2M & 13.61 & 0.5756 & 4.2M \\  
  V-LSTM+PtrNet & 21.36 & 0.5934 & 66.9M & 35.62 & 0.6564 & 54.5M & 17.95 & 0.5620 & 43.6M & 13.13 & 0.5703 & 46.5M \\
  ATTOrderNet & 25.73 & 0.6122 & 40.9M & 40.00 & 0.6763 & 32.6M & \textbf{29.17} & 0.6200 & 13.4M & 13.57 & 0.5733 & 16.3M \\
  SE-Graph & \textbf{26.73} & \textbf{0.6251} & 12.4M & \textbf{45.97} & \textbf{0.6995} & 11.4M & 27.55 & \textbf{0.6282} & 4.5M & \textbf{15.07} & \textbf{0.5838} & 11.6M \\

\end{tabular}}

\setlength{\tabcolsep}{1.5mm}{
\begin{tabular}{ccccccccccccc}
\midrule
\multirow{2}{1.5cm}{\centering Paragraph Level} & \multicolumn{3}{c}{Press}                                     & \multicolumn{3}{c}{Statements}                                  & \multicolumn{3}{c}{Journal} 
                            & \multicolumn{3}{c}{Lyrics} \\ \cline{2-13}
                         & pmr & wlcs-l & \#pm & pmr & wlcs-l & \#pm & pmr & wlcs-l & \#pm & pmr & wlcs-l & \#pm \\ \midrule
LSTM+PtrNet & 42.33 & 0.6543 & 10.0M & 43.12 & 0.6582 & 6.6M & 5.24 & 0.4549 & 11.2M & 0.26 & 0.1765 & 8.9M \\  
  V-LSTM-PtrNet & 35.04 & 0.6304 & 64.1M & 33.01 & 0.6346 & 53.7M & 4.85 & 0.4424 & 67.6M & \textbf{0.66} & \textbf{0.1891} & 60.7M \\
  ATTOrderNet & 42.57 & 0.6555 & 55.4M & \textbf{45.79} & 0.6692 & 45.0M & 5.27 & 0.4595 & 41.6M & 0.49 & 0.1751 & 52.1M \\
  SE-Graph & \textbf{46.90} & \textbf{0.6749} & 13.3M & 42.66 & \textbf{0.6704} & 12.0M & \textbf{5.29} & \textbf{0.4618} & 20.02M & 0 & 0.1719 & 12.4M \\
\bottomrule
\end{tabular}}
%\end{tabularx}
\caption{Main results of sequence ordering task at the paragraph and the sentence level, where \#pm refers to the number of model parameters, V-LSTM+PtrNet stands for Variant-LSTM+PtrNet; For LSTM+PtrNet and Variant-LSTM+PtrNet, since there is no publicly available source code, we base our experiment on our own replication.}
\label{tab:basic-task-performance-stats}
\end{center}
\end{table*}

\subsection{Metrics}

We measure the ranking performance[sorting performance] from two aspects.
From a global perspective, we use Perfect Match Ratio (\textbf{PMR}) and Accuracy (\textbf{Acc}) as the proxy for the absolute position measurement; from a local perspective, Kendall's tau(\textbf{$\tau$}) and Weighted Longest Common Subsequence (\textbf{WLCS-l}) are used as pair-wise position metrics.
\paragraph{Perfect Match Ratio(PMR)}is the most stringent measurement in this task. It calculates the ratio of exactly matching orders: $PMR= \frac{1}{K}\Sigma^{K}_{i=1}\mathbbm{I}(\hat{o}^{i}=o^{i\ast})$ , where $\hat{o}^{i}$ and $o^{i\ast}$ are predicted and correct orders of the i-th text respectively. 
\paragraph{Accuracy(Acc)} is a measure of how often the absolute position of a sentence was correctly predicted. Compared with PMR, this is a finer metric to measure how well the model performs on finding the absolute position.
\paragraph{Kendall's tau($\tau$)}  is one of the most frequently used metrics for the automatic evaluation of document coherence. It could be formalized as: $\tau = 1 - 2 \times (number\ of\  inversions) / \binom{n}{2} $, where n is the length of the sequence and the number of inversions denotes the number of pairs in the predicted sequence with incorrect relative order. This metric ranges from -1 (the worst) to 1 (the best).
% ###############################################################################
\paragraph{Length Adapted Weighted Longest Common Subsequence(WLCS-l)} is an adapted version of original ROUGE-w, the metric of the extent of sequence overlapping~\cite{lin-2004-rouge}. 
%Compared with other LCS-based metrics in the ROUGE family, A-ROUGE-w encourages the model to generate successive sequence when the overlapping proportion is same. 
Compared with other LCS-based metrics in the ROUGE family, WLCS-l encourages the model to generate successive sequence when the overlapping proportion is same while automatically decreasing with the total sequence length.
It could be formalized as: 
\begin{align}
    P_{wlcsl} &= f^{-1}\Bigg(\frac{WLCS(\hat{O},O)}{f(n)}\Bigg) \\
    R_{wlcsl} &= f^{-1}\Bigg(\frac{WLCS(\hat{O},O)}{f^{2}(n)}\Bigg) \\
    F_{wlcsl} &= \frac{(1+\alpha^{2})R_{wlcsl}P_{wlcsl}}{R_{wlcsl}+\alpha^2P_{wlcsl}}
\end{align}
where $n$ represents the length of passage. $\hat{O}$ and $O$ represent the sequence of predicted order and truth order sequence, respectively. The two functions $f$ and $WLCS$ are the same in ~\cite{lin-2004-rouge}.
Intuitively, we hope the model generates consecutive suborder instead of segments of skip-grams. We use the package py-rouge\footnote{\url{https://pypi.org/project/py-rouge/}} to calculate the WLCS-l score with the default alpha value being adopted.
% ##############################################################################

\subsection{Within and Cross-domain Performance}

To generally examine the performance of each framework, we first run models on all datasets.
Table~\ref{tab:basic-task-performance-stats} shows the test results of established approaches in two types of tasks.
For each dataset, we report the PMR, WLCS-l and number of parameters of each measure. 
Note that PMR values might be lower than those in previous works since we remove all the texts containing only one sequence. 
The results confirm our expectation that graph model is much more powerful for ordering task, since it automatically learns directional information flow among each sequence rather than through a central weight matrix as in ATTOrderNet or by a single hidden layer as in Variant-LSTM+PtrNet. 
At the sentence level, SE-Graph almost dominates all previous models on both the global coherence (measured by PMR) and the local coherence(measured by WLCS-l). 
At the paragraph level, SE-Graph continues to be the state-of-the-art method in most cases. It acquires the best WLCS-l scores on three datasets and the best PMR on two, which indicates its adaptability on long sequence tasks. 
% The following of this section will articulate the discovery with cross-domain experiments.

\begin{table*}[ht]
\centering
\begin{subtable}{0.35\linewidth}
\centering
\setlength{\tabcolsep}{0.4mm}{
\begin{tabular}{ccccc}
      \toprule
      \multirow{2}{*}{Train}&\multirow{2}{*}{Model}&\multicolumn{3}{c}{Test}\\
      \cline{3-5}
      & & arXiv & AAN & NIPS\\
      \midrule
      \multirow{4}{*}{arXiv} 
      & LSTM+PtrNet & 0.73 & 0.77 & 0.81$^{\triangle}$\\
      & V-LSTM+PtrNet & 0.77 & 0.80 & \textbf{0.83}$^{\triangle}$\\
      & ATTOrderNet & 0.75 & 0.79 & 0.81$^{\triangle}$\\
      & SE-Graph & \underline{\textbf{0.78}} & 0.80 & 0.80$^{\triangle}$\\
    \midrule
    \multirow{4}{*}{AAN} 
    & LSTM+PtrNet & 0.64 & 0.74 & 0.65\\
      & V-LSTM+PtrNet & 0.73 & \underline{\textbf{0.81}} & 0.75\\
      & ATTOrderNet & 0.64 & 0.76 & 0.65\\
      & SE-Graph & 0.72 & \underline{\textbf{0.81}} & 0.80$^{\triangle}$\\
    \midrule
    \multirow{4}{*}{NIPS} 
    & LSTM+PtrNet & 0.62 & 0.65 & 0.70\\
      & V-LSTM+PtrNet & 0.64 & 0.64 & 0.70\\
      & ATTOrderNet & 0.66 & 0.70 & 0.73\\
      & SE-Graph & 0.66 & 0.71 & \underline{0.75}\\
    \bottomrule
\end{tabular}}
\caption{Results of \textbf{$\tau$} on the sentence ordering task.}
\label{tab:sent-cross-domain-kendall}
\end{subtable}%
\begin{subtable}{0.68\linewidth}
\centering
\setlength{\tabcolsep}{0.3mm}{
\begin{tabular}{ccccc}

      \toprule
      \multirow{2}{*}{Train}&\multirow{2}{*}{Model}&\multicolumn{3}{c}{Test at paragraph level / Test at sentence level}\\
      \cline{3-5}
      & & News & Statements & Economist\\
      \midrule
      \multirow{4}{*}{News} 
      & LSTM+PtrNet & 69.33 / 76.04 & 28.42 / 40.81 & 9.68 / 26.77\\
      & V-LSTM+PtrNet & 61.48 / 74.55 & 25.26 / 40.64 & 4.77 / 25.62\\
      & ATTOrderNet & \underline{\textbf{70.19}} / \underline{\textbf{77.78}} & 39.37$^{\diamondsuit}$/ 44.82 & 13.64$^{\diamondsuit}$/ 26.90$^{\diamondsuit}$\\
      & SE-Graph & 66.22 / 69.80 & 31.72 / 45.42$^{\diamondsuit}$ & 12.69 / 24.29\\
      \midrule
    \multirow{4}{*}{Statements} 
    & LSTM+PtrNet & 24.96$^{\diamondsuit}$/ 49.98 & 61.66 / \underline{\textbf{60.14}} & 14.50 / 29.44\\
      & V-LSTM+PtrNet & 17.86 / 51.12 & 52.67 / 59.19 & 8.55 /  27.44\\
      & ATTOrderNet & 23.54 / 52.77$^{\diamondsuit}$ & \underline{\textbf{63.41}} / 59.83 & 16.11$^{\diamondsuit}$/ 30.06$^{\diamondsuit}$\\
      & SE-Graph & 22.71 / 39.84 & 48.50 / 55.68 & 15.88 / 27.35\\
      \midrule
    \multirow{4}{*}{Economist} 
    & LSTM+PtrNet & 25.74$^{\diamondsuit}$/ 52.08 & 36.96$^{\diamondsuit}$/ 50.81$^{\diamondsuit}$ & 37.42 / 45.64\\
      & V-LSTM+PtrNet & 18.75 / 45.03 & 39.67 / 44.69 & 31.44 /  39.75\\
      & ATTOrderNet & 23.60 / 52.64$^{\diamondsuit}$ & 36.15 / 49.55 & \underline{\textbf{37.80}} / \underline{\textbf{46.71}}\\
      & SE-Graph & 21.78 / 38.61 & 20.14 / 40.16 & 32.69 / 36.28\\
      \bottomrule
\end{tabular}}
\caption{Results of \textbf{Acc} on both text ordering tasks on our courpus.}
\label{tab:text-cross-domain-acc}
\end{subtable}%
\caption{In the above, tables V-LSTM+PtrNet stands for Variant-LSTM+PtrNet. 
% For experiments in the abstracts, We stress the best performance in each column with boldface, underline the best performance within each domain and attach an empty triangle superscript to cross-domain results that exceeds the highest performance within the original domain. For experiments in our corpus, We stress the best performance in each column with boldface, underline the best performance within each domain and attach an empty diamond superscript to cross-domain results that ranks the top in each cell.
We stress the best performance in each column with boldface and underline the best performance within each domain. For experiments in the abstracts, we attach an empty triangle superscript to cross-domain results that exceeds the highest performance within the original domain. For experiments in our corpus, we attach an empty diamond superscript to cross-domain results that ranks the top in each cell.}
\end{table*}

% \caption{For document level task, where V-LSTM+PtrNet stands for Variant-LSTM+PtrNet. We stress the best performance in each column with boldface, underline the best performance within each domain and attach an empty diamond superscript to cross-domain results that ranks the top among four. Variant-LSTM+PtrNet and SE-Graph illustrate strong capacity on pair-wise ordering while ATTOrderNet and LSTM+PtrNet tends to show their strength in absolute position finding.}

We examine the domain adaptability of available approaches by first training and tuning the model on data from one domain, and then test them on test set from another. This procedure mimics the practice in reality - when there is no gold passage corpus for training, one must utilize data from another resource then go through the pretrain-finetune process. 
% Table~\ref{tab:sent-cross-domain-kendall}-\ref{tab:text-cross-domain-acc} show results measured by $\tau$ and some results measured by ACC over two task levels. 

Table~\ref{tab:sent-cross-domain-kendall} shows the results of cross-domain experiments on academic abstract datasets, which supports our assertion that the prevalent researches are limited on specific text genre. When measured by $\tau$, all of the four evaluated models are capable of transferring the knowledge learning from arXiv to NIPS. This conclusion can also be reinforced when measured by Acc. Nearly all triangles locate within the complex row of arXiv, indicating that it could be a general language resource of academic abstract for sentence ordering, in terms of pair-wise order.
% Varient-PtrNet acquires the best performance reaching 0.83, significantly higher than the SOTA when trained within NIPS itself. In addition, SE-Graph and Varient-PtrNet both meets 0.80 when test on AAN under a cross-domain setting, approximating the 0.81 SOTA when learned within AAN itself.
% For absolute coherence measure by Acc, the trend is much more salient. All boldface figures and nearly all triangles locate within the complex row of arXiv, indicating that it could be a general language resource of academic abstract for sentence ordering in terms of absolute position. 
% Moreover, ATTOrder is proved to be outstanding on absolute position capturing since most two thirds of boldface figures and underline figures locates in its single row. 

Table~\ref{tab:text-cross-domain-acc} gives a glance of model performance measured by Accuracy on various domains. 
It is quite obvious that our data resources suffer from less overlapping, which makes it valuable to be used for a training and testing purposes in a broader and more realistic sense.
%which makes it valuable to be used for the training and test purposes in a broader and more realistic sense. 
Unlike corpus for the sentence-level task, results show models only work when training on the corresponding domain. 
This conclusion is also valid when measured by $\tau$. 
In addition, in order to test the effect of domain on sentence ordering tasks, we adopt each paragraph in the corpus as a whole passage and test sentence ordering capability of evaluated models.
Results also show that Economist is a hard dataset for ordering tasks both at the paragraph and at the sentence level. This is perhaps out of its argument nature, where sentences and paragraphs are connected more by higher level semantic relations instead of hint words. 
% From the angle of relative coherence, we can find SE-Graph receives most credits in both within-domain and cross-domain task. 

When measured by Accuary, ATTOrderNet dominates all other models by acquiring the state-of-the-art results in all in-domain and nearly all cross-domain tests in our corpus, which confirms capacity of self-attention structure on catching the absolute coherence.  Table~\ref{tab:text-cross-domain-acc} illustrates the performance of ATTOrderNet at both the sentence level and the paragraph level. This conclusion can also be applied to the ordering task on academic abstracts. In addition, SE-Graph distinguishes itself when measured by $\tau$. Table~\ref{tab:sent-cross-domain-kendall} reveals a comparison of the four models on academic abstract corpus. SE-Graph gains state-of-the-art performance in all the within domain in-domain experiments and competitive results with the benchmark when tested under a cross domain cross-domain circumstances. The conclusion is also well supported by results at the paragraph-level task.

It is easier to transfer knowledge from Economist to News and Statements Corpus at the sentence level than at the paragraph level. Benchmark accuracy for cross-domain inference reaches over 0.5 in when transfer the knowledge learned from Economist to News and Statements. However, when we try to transfer the knowledge in the same way at the paragraph level, the benchmark performance collapse on News. This phenomenon could be a result of less sentences but more paragraphs in News, which makes the sentence-level task easier.
\subsection{Learnability and Robustness Analysis}
We picked out arXiv and Economist as the representative language source on behalf of general discourses at the sentence level and the paragraph level respectively since the results of cross-domain experiments in the last subsection has illustrated their comprehensiveness. To mimic practical environment we create mini datasets and noisy datasets as approaches to test model performance.

We report WLCS-l instead of $\tau$ as a proxy for local coherence. The motivation behind this is a common phenomenon: misplaced successive sentences are easier to be recap than sequences round its absolute position connected in a skip-gram manner with its supposed siblings. We observe this phenomenon from the human evaluation. Sequences carry consecutive meaning in one clique tends to give human smoother reading experience, which can be perfectly measured by WLCS-l. Besides, Acc is set as the indicator of absolute coherence.

In the following parts, we first introduce the artificial datasets we made and then analyze the experiment results.

\begin{figure*}[htbp]
\centering
\begin{subfigure}[t]{0.25\linewidth}
\centering
\includegraphics[width=1.5in, height=1.2in]{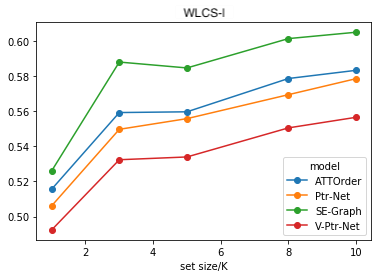}
\caption{Results of WLCS-w on arXiv}
\label{fig:scarce_experiments_arxiv_wlcsl}
\end{subfigure}%
\begin{subfigure}[t]{0.25\linewidth}
\centering
\includegraphics[width=1.5in, height=1.2in]{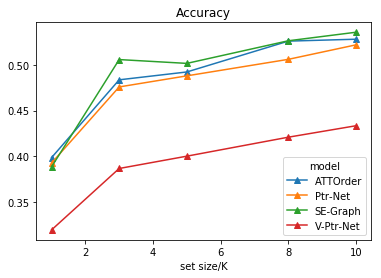}
\caption{Results of Acc on arXiv}
\label{fig:scarce_experiments_arxiv_acc}
\end{subfigure}%
\begin{subfigure}[t]{0.27\linewidth}
\centering
\includegraphics[width=1.5in, height=1.2in]{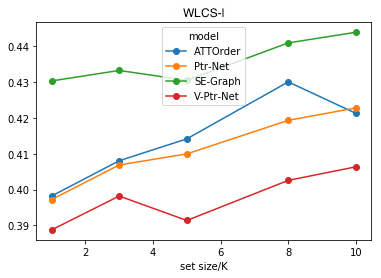}
\caption{Results of WLCS-l on Economist}
\label{fig:scarce_experiments_journal_wlcsl}
\end{subfigure}%
\begin{subfigure}[t]{0.25\linewidth}
\centering
\includegraphics[width=1.5in, height=1.2in]{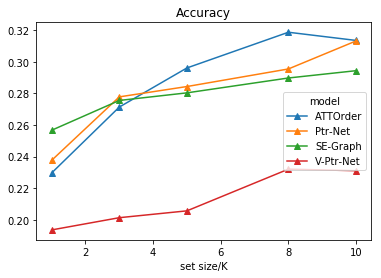}
\caption{Results of Acc on Economist}
\label{fig:scarce_experiments_journal_acc}
\end{subfigure}%
\centering
\caption{Some metrics on mini datasets, where ATTOrder represents ATTOrderNet; Prt-Net represents LSTM+PtrNet, where V-Ptr-Net stands for Variant-LSTM+PtrNet.}
\label{fig:minidata}
\end{figure*}

\begin{figure*}[htbp]
\centering
\begin{subfigure}[t]{0.25\linewidth}
\centering
\includegraphics[width=1.5in, height=1.2in]{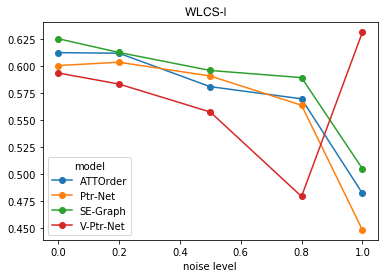}
\caption{Results of WLCS-l on arXiv}
\label{fig:noisy_experiments_arxiv_wlcsl}
\end{subfigure}%
\begin{subfigure}[t]{0.25\linewidth}
\centering
\includegraphics[width=1.5in, height=1.2in]{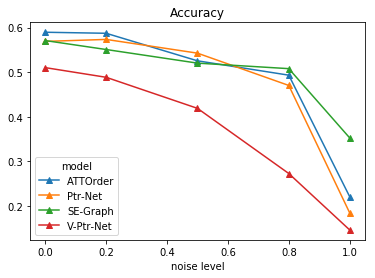}
\caption{Results of Acc on arXiv}
\label{fig:noisy_experiments_arxiv_acc}
\end{subfigure}%
\begin{subfigure}[t]{0.27\linewidth}
\centering
\includegraphics[width=1.5in, height=1.2in]{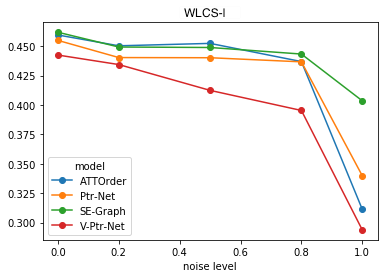}
\caption{Results of WLCS-l on Economist}
\label{fig:noisy_experiments_journal_wlcsl}
\end{subfigure}%
\begin{subfigure}[t]{0.25\linewidth}
\centering
\includegraphics[width=1.5in, height=1.2in]{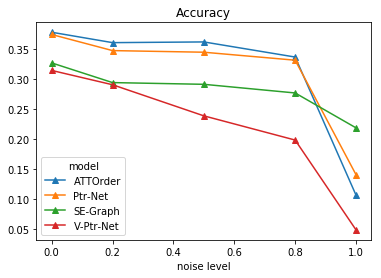}
\caption{Results of Acc on Economist}
\label{fig:noisy_experiments_journal_acc}
\end{subfigure}%
\centering
\caption{Some metrics on noisy sources, where the legend remains the same as those in Figure 1.}
\label{fig:noisydata}
\end{figure*}

\subsubsection{Mini and Noisy Datasets}
Aiming at studying the learnability of models under low-resource data scenarios, we sample mini dataset to mimic the low-resource occasion. 
To be more precisely, we sample children training and dev set  as mini datasets from the parent corpus in a bootstrap manner by randomly picking out a certain number of instances from the parent set iteratively and independently while fixing the proportion between train and dev.
We divide the scale of children training set into five levels -- 1k, 3k, 5k, 8k, 10k.
In experiment, we run the models on these mini datasets and observe the variation of our metrics.
%, with the corresponding size for test and dev set of 50, 150, 250, 400, 500. 
% Considering the quality and comprehensiveness, we select arXiv and Journal as our parent datasets, which serve as representative of sentence level corpora and text level corpora respectively. 

% Section~\ref{sec:experiments} will show details about the implementation.
In an attempt to find the robustness of available methods, we generate the noisy datasets for noisy experiment. 
To be more specific, we add "noise" to a specific sequence by randomly adopting one approach from the following three options: 

\begin{itemize}
    \item \textbf{Insert}. We create the "Insert" noise by randomly picking an ad from a list\footnote{We retrieve the ad slogans from two website.
    In total, there are 100 pieces of ads in the set \url{https://marketingwit.com/famous-advertising-slogans},\url{https://www.thebalancecareers.com/best-advertising-taglines-ever-39208}} and insert it in front of the sequence processed.
    The intuition behind this approach is that ads usually mixed in the middle of an article and hard to remove. 
    % Although writing a regular expression filter or recognizing the slogan with certain manually created pattern may alleviate this issue, completely remove them is not simple even in industry.
    As a result, we try to simulate this occasion by a inverse measure - "Insert" ads into the passage.
    
    \item \textbf{Remove}. We pop out the sequence from the complete text under the "Remove" noise mode.
    The motivation behind is to mock the scenario where processing PDF with existing parsing package may face a format mismatch, causing content missing. Thus, we simply "Remove" some sequences to mimic this scenario. 
    
    \item \textbf{Modify}. We randomly pick 50\% of all the tokens in one sequence and modify them in three possible ways: concatenating, splitting and replacing. 
    For concatenating, we concatenate the current token with another token right behind if the current token does not locates at the tail position. 
    For splitting, we split the current token in a randomly selected position when its length is not smaller than two. 
    For replacing, we randomly replace one of the letters in the token with one character in the substitution list\footnote{The list contains all possible numbers, letters and symbols that is available in an English context}. 
    
\end{itemize}

In order to uniformly permute our noise in the whole corpus, for each sequence, we assign an probability $p$ of being "contaminated", which means each sentence has an equal chance to be polluted by "noise". 
We define the noisy level of dataset with the parameter $p$. 
In our experiment, we verify these models on four levels of $p$: 20\%, 50\%, 80\%, 100\%. 
This measure is to emulate that text harvested from a scanner or a photo converter that may contain some mistakes in word spelling out of misrecognizing letter from the picture.

\subsubsection{Experiments on Learnability}
To effectively examine learnability, we train our models on mini datasets. In order to acquire accurate measures, we conduct 20 and 10 bootstrap tests on mini datasets whose training size is 1K and 3K, and 5 bootstrap tests on mini datastes whose training set is 5K, 8K and 10K respectively. 

Figure~\ref{fig:scarce_experiments_arxiv_wlcsl} and ~\ref{fig:scarce_experiments_journal_wlcsl} show that in general, SE-Graph outperforms other methods with a high margin if measured by wlcs-l, and often follows the trends of ATTOrderNet in both the sentence and the paragraph levels.
% , probably because the message passage mechanism takes advantage of the graph building supported by dependency parsing in its prepossessing, where sentence nodes were connected if they share common entities. 

In terms of Accuracy, Figure~\ref{fig:scarce_experiments_arxiv_acc} illustrates that except Variant-LSTM+Ptr-Net, all three models share relatively similar performance under a data-scarce environment at the sentence level. At the paragraph level, although self-attention-based method dominates all others when measured by Accuracy, it fails to keep its superiority in an extreme occasion when training size is less than 3K, as the cross point in Figure~\ref{fig:scarce_experiments_journal_acc} indicates. On the country, GRN structure reveals its strong adaptability in an extreme low-resource environment. 

\subsubsection{Experiments on Robustness}
To effectively examine robustness, we train our models on artificially sampled noisy datasets.
 Figure~\ref{fig:noisy_experiments_arxiv_wlcsl}-\ref{fig:noisy_experiments_arxiv_acc} illustrate the results at the sentence level. SE-Graph illustrates the strongest noise resistance against the other three.
%  In contrast, Variant-PtrNet remains the worst in all cases before the noise level goes reaches to above 80\%. 
 This suggests the robustness of SE-Graph at the sentence level from a relative coherence angle. For global coherence, SE-Graph is supposed to be the most insensitive. Although it remains slightly lower accuracy under a light-noisy condition, its performance enjoys the least decrease when noise level raise to an extreme level.

Figure~\ref{fig:noisy_experiments_journal_wlcsl}-\ref{fig:noisy_experiments_journal_acc} plots the results at the paragraph level. The WLCS-l curves indicate the insensitivity of graph framework. Among the four models, only SE-Graph keeps its WLCS-l above 0.4 when noise level comes to 100\% whereas the metrics of all other structures drop below 0.35. For absolute positions, the modest slope of green curve reinforces this conclusion, although ATTOrderNet dominates in terms of Accuracy under low noisy circumstances, the Accuracy measure of SE-Graph suffers from the least drop rate when noise reaches to the highest level.

The outstanding noise resistance and low-resource favoring property of the SE-Graph probably lie in its graph building procedure. When the noise level rises or sample size shrinks, LSTM sentence encoder structure has a hard time to encode sequences precisely. While all other models simply rely on the encoded sequence representation from LSTM, SE-Graph explicitly utilizes the connection from dependency parsing. Thus, as long as the entities in the sequence are not hurt, SE-Graph could ensure the right topographical structure of information flow in the successive graph encoding process. 
As a result, for document ordering, where much more entities are in one sequence in average, there is a higher chance for arbitrary two sentences or paragraphs to remain correctly connected with each other.

\begin{figure*}[htbp]
\centering
\begin{subfigure}[t]{0.3\linewidth}
\centering
\includegraphics[width=1.8in, height=1.44in]{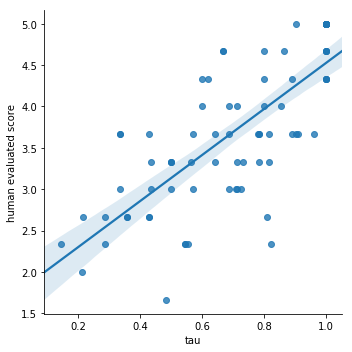}
\caption{$\tau$ does not correlates well with human judgements for passage coherence.}
\label{fig:correlation_experiments_arxiv_tau}
\hspace{3mm}
\end{subfigure}%
\hspace{3mm}
\begin{subfigure}[t]{0.3\linewidth}
\centering
\includegraphics[width=1.8in, height=1.44in]{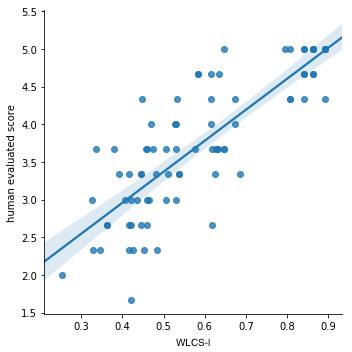}
\caption{WLCS-l correlates well with human judgements for passage coherence.}
\label{fig:correlation_experiments_arxiv_acc}
\end{subfigure}%
\hspace{3mm}
\begin{subfigure}[t]{0.3\linewidth}
\centering
\includegraphics[width=1.8in, height=1.44in]{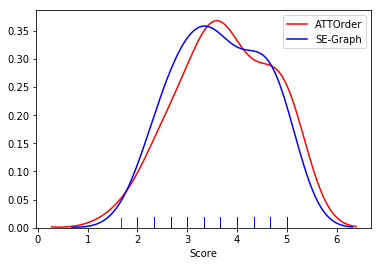}
\caption{ATTOrderNet slightly outperforms SE-Graph in human evaluation.}
\label{fig:dist_rating}
\end{subfigure}%
\centering
\caption{Correlations between human rating and $\tau$ and WLCS-l respectively, in addition with human rating distribution for inference performance of two models. }
\label{fig:human eval}
\end{figure*}

\subsection{Human Evaluation}
In our experiments, we randomly sampled 40 passages from the test set of arXiv abstract and pick the predicted articles of ATTOrderNet and SE-Graph. 
As a result, in total, 80 passages are evaluated by human judges. To avoid the fatigue effect brought by long-time evaluating, we half the predicted orders from each model into two sets.
Thus, passage 1-20 from each model are put into one package while the other 40 passages are put into another.

We distribute the packages among a total of 6 human judges (3 judges per package). Thus, each judge is presented with 40 rearranged permutation of sentences in total from two models. 
They were asked to assign a single coherence rating for each passage permutation. 
The ratings were on a scale of 1 to 5, with 1 being very incoherent and 5 being perfectly coherent. 
The standard is listed as the following:
\begin{itemize}
    \item 1 - Nonsense: What it says makes no sense.
    \item 2 - Wired: It looks rigid but one can understand some segments.
    \item 3 - Acceptable: One can understand the meaning but it's not an easy task.
    \item 4 - Coherent: One can understand the meaning with no difficulty although there are some flaws.
    \item 5 - Fluent: One can understand the meaning and he cannot tell whether it is from human or machine.
\end{itemize}
We do not provide any additional instructions or examples of scale as we wanted to capture the intuitive idea of coherence from our judges.

% \begin{table}[!h]
% \begin{center}
% \begin{tabular}{p{20pt}p{180pt}}
% \toprule
% \textbf{Score} & \textbf{Standard Description}\\ \midrule
%     1 & {\it{Nonsense. What it says makes no sense.}} \\ \midrule
%     2 & {\it{Wired. It looks rigid but I can understand some segments}} \\ \midrule
%     3 & {\it{Acceptable. I can understand the meaning but it's not an easy task.}} \\ \midrule
%     4 & {\it{Coherent. I can understand the meaning with no difficulty although there are some flaws.}} \\
%     \midrule
%     5 & {\it{Fluent. I can understand the meaning and I cannot tell whether it is from human or machine.}} \\
% \bottomrule
% \end{tabular}
% \caption{Scoring standard for human raters to evaluate the rearrangements of the abstract permutations.}
% \label{tab:demo-standard}
% \end{center}
% \end{table}

We compute the inter-rater agreement by using Pearson's correlation analysis. We concatenate the ratings for each half of the samples from one model into one array. In this way, we obtain three arrays simulating score from three different raters. We correlate the ratings given by each judge with the average ratings given by these judges. For interrater agreement we report the average of 3 such correlations which is 0.85 (p-value = 1.2e-23). Krippendorff's $\alpha$~\cite{artstein-poesio-2008-survey} can be used for inter-rater agreement with interval scales like the one we have. In our case, the $\alpha$ values were 0.53. The moderate value of alpha indicates that judging coherence of a passage is indeed a difficult task, especially when detailed instructions or examples of scales are not given.
%~\cite{gandhe-traum-2008-evaluation}.

In order to assess which metrics is the best automatic measure of sentence ordering coherence, we regress the average ratings by human judges on $\tau$, Accuracy and WLCS-l. Although the the regression on $\tau$ reflects the effectiveness of the wildly used metrics with an R-square of 0.571, the regression on Accuracy and WLCS-l indicates 
% the inferiority of $\tau$ 
% when it's used to measure human experience
their appropriateness with R-square of 0.618 and 0.665, respectively. The results proves that WLCS-l, which has never been considered as one way to measure the validity of sentence ordering methods is indeed correlates the best with human judgements. 
% ############################################################
Note that our metric differs from the original ROUGE-w by doubling the score operator for the denominator. In this way, the  $F_{wlcsl}$ would adjusted itself with the length of texts - the score will automatically diminish when the text goes longer. This stimulates human reaction to long texts - people are much more likely to get lost in longer (and usually more complex) texts and thus a highly coherent permutation is needed as a compensation. Fig~\ref{fig:correlation_experiments_arxiv_tau}-\ref{fig:correlation_experiments_arxiv_acc} shows the plot when regress human rates on $\tau$ and WLCS-l. 
% ############################################################

% \begin{figure}[!h]
% \centering
% \begin{subfigure}[t]{1.0\linewidth}
% \centering
% \includegraphics[width=2.5in, height=2.0in]{regress_tau.png}
% \caption{$\tau$ does not correlates well with human judgements for passage coherence.}
% \label{fig:correlation_experiments_arxiv_tau}
% \end{subfigure}%

% \vspace{7mm}
% \begin{subfigure}[t]{1.0\linewidth}
% \centering
% \includegraphics[width=2.5in, height=2.0in]{regress_rouge.png}
% \caption{ROUGE-w correlates well with human judgements for passage coherence.}
% \label{fig:correlation_experiments_arxiv_acc}
% \end{subfigure}%
% \centering
% \caption{single coherence rating per permutation}
% \end{figure}

To compare the real human experience of machine planned abstracts, we plot the distribution curves of over all human rating of each model. SE-Graph is proved to be slightly worse compared with ATTOrderNet. The scores for SE-Graph ordered passages peaks at roughly 3 while that for ATTOrderNet peaks at 3.3. This is counter-intuitive since we know from the above analysis that a higher wlcs-l score tends to correlates with a better human rating. However, we observe a phenomenon that rating is sometimes affected by the intrinsic complexity of the passage. Thus, when readers finished reading the passages generated by graph-based method, they then will have a basic impression of the articles, which leads to shorter time spent on understanding predicted text of self-attention and thus causes an illusion of its better performance among raters. In fact, 4 out of 6 annotators rated output of SE-Graph before they score the alternative file, and they all express the impact of intrinsic complexity on coherence rating.

\section{Conclusion}
\label{sec:conclusion}

In this work, we propose the paragraph ordering task as a supplement of sentence ordering, and 4 datasets with diverse domain and low level of overlapping. Based on the public data source and corpus we presented, we evaluated four structures for text ordering by conducting experiments within and cross different domains on both the sentence level and the paragraph level. We use these multi-genre benchmarks to show the efficiency of SE-Graph, the SOTA approach for sentence ordering based on graphical neural network structure. We also conduct examination on the learnability and robustness of the existing methods by randomly downsampling passages from the parent dataset to mimic the small scale data condition, and by artificially blending noise into the original dataset to imitate a real-world scenario, which help us confirm the excellent learnability and strong noise resistance of the SOTA approach. We also carry out a human evaluation between attention-based and graph-based models, whose results strongly support that WLCS-l, a metric we adopt to measure text ordering for the first time, exhibits a higher correlation with human score than $\tau$. 
This conclusion provides a hint for future research, which is expected to focus more on improving common sequence instead of concentrating on the order between sequence pairs. Flexible decoder generator may be a direction with greater value, since it may allow a machine to generate sentence in an order that is easier to form coherence clique compared with starting rigidly with the head sequence.
% \citet{barzilay-lapata-2008-modeling}
\section*{Acknowledgments
}
We thank the anonymous reviewers for their helpful comments on this paper. This work
was partially supported by  National Key Research and Development Project (2019YFB1704002) and National Natural
Science Foundation of China (61876009 and 61572049). The corresponding author of this
paper is Sujian Li.

\section*{Bibliographical References}
\label{main:ref}

\bibliographystyle{lrec}
\bibliography{lrec2020W-xample}

\end{document}